\documentclass[letterpaper, 10 pt, conference]{ieeeconf}  
\usepackage{amsmath} 
\usepackage{amssymb}  
\usepackage{graphicx}
\usepackage{epstopdf}
\usepackage{amsmath}
\usepackage{mathtools}
\usepackage{dsfont}

\newcommand{\beq}{\begin{equation}}
\newcommand{\eeq}{\end{equation}}
\newcommand{\rr}{{\mathbb R}}
\newcommand{\zz}{{\mathbb Z}}

\newcommand{\argminF}{\mathop{\mathrm{argmin}}\limits}   

\newcounter{algorithmctr}[section]
\renewcommand{\thealgorithmctr}{\thesection.\arabic{algorithmctr}}
{\refstepcounter{algorithmctr}\begin{list}{}{%
\setlength{\rightmargin}{0\linewidth}%
\setlength{\leftmargin}{.05\linewidth}
\setlength{\itemsep}{1pt}
\setlength{\parskip}{0pt}
\setlength{\parsep}{0pt}}%
\rmfamily\small
\item[]{\setlength{\parskip}{0ex}\hrulefill\par%
\nopagebreak{\bfseries\textsf{Algorithm \thealgorithmctr~}}}}%
{{\setlength{\parskip}{-1ex}\nopagebreak\par\hrulefill} \end{list}}
\IEEEoverridecommandlockouts
\newtheorem{assumption}{Assumption}

\newtheorem{remark}{Remark}

\title{\LARGE \bf
Autonomous Racing using Learning Model Predictive Control
}

\author{Ugo Rosolia, Ashwin Carvalho and Francesco Borrelli
\thanks{Ugo Rosolia, Ashwin Carvalho and Francesco Borrelli are with the Department of Mechanical Engineering, University of California at Berkeley ,
Berkeley, CA 94701, USA
{\tt\small \{ugo.rosolia,  ashwinc, fborrelli\} $@$ berkeley.edu}}%
}

\begin{document}

\maketitle
\thispagestyle{empty}
\pagestyle{empty}

\begin{abstract}
A novel learning Model Predictive Control technique is applied to the autonomous racing problem. 
The goal of the controller is to minimize the time to complete a lap.
The proposed control strategy uses the data from previous laps to improve its performance while satisfying safety requirements.
A  system identification technique is proposed to estimate the vehicle dynamics.
Simulation results with the high fidelity simulator software CarSim show the effectiveness of the proposed control scheme.
\end{abstract}

\section{INTRODUCTION}
Nonlinear model predictive control is an appealing technique for autonomous driving because of its ability to handle input and state constraints as well as nonlinearities introduced by the vehicle dynamics. Recently, the effectiveness and the real-time feasibility of this control strategy has been demonstrated in \cite{MPC2,MPC3}. However, real-time path generation for racing applications in real-world conditions remains challenging. If the objective is the minimization of the lap time, the controller has to plan the trajectory on a sufficiently long time interval to avoid aggressive maneuvers that would lead the vehicle outside the track. For instance, in a straight line before a curve the controller should limit the velocity to allow turning.
Furthermore, when solving the problem with a receding horizon approach, the resulting closed-loop trajectory is not guaranteed to be optimal for the original problem.

In \cite{RacingETH} two approaches are presented for autonomous racing. In the first one, a high level MPC controller is used to generate a feasible trajectory for the low level tracking MPC. In the second approach, a single layer Model Predictive Contouring Control (MPCC) is presented, where the controller objective is a trade-off between the progress along the track and the contouring error. Also in \cite{RacingLMS}, the authors compared two approaches, the first one based on a tracking MPC, and, the second one, based on a MPC that aims to minimize the lap time. The authors pointed out the importance of the horizon length, which is necessary to reach good performance. In \cite{kapania2015path} the authors proposed an iterative learning control (ILC) approach for autonomous racing. The proposed ILC tracks an aggressive trajectory computed off-line using the techniques proposed in \cite{theodosis2011generating}. The authors showed the effectiveness of the proposed ILC with experimental testing on a full size vehicle.

In this paper we propose to tackle the minimum lap time problem as reference free iterative control problem using the novel technique presented in \cite{ILMPC}.
At each $j$-th iteration the controller starts from the same starting point and it has to minimize the traveling time to cross the finish line. Under no model mismatch, the Learning MPC (LMPC) presented in \cite{ILMPC} guaranties that \emph{(i):} the $j$-th iteration's  cost does not increase compared to the $j-1$-th iteration  cost (non-increasing cost at each iteration), \emph{(ii):} state and input constraints are satisfied at iteration $j$ if they were satisfied at iteration $j-1$ (recursive feasibility), \emph{(iii):} if the controller goal is to regulate the system to an equilibrium point $x_F$, then such an equilibrium point is asymptotically stable, \emph{(iv)}: if system converges to a steady state trajectory as  the number of iterations $j$ goes to infinity, then such a trajectory is locally optimal. The assumption of no model mismatch is not satisfied in the racing application especially due to the highly dynamic range of maneuvers.
Therefore, in this work we propose to couple the LMPC with a 
system identification algorithm and we show, through high fidelity simulations with CarSim, that the controller successfully improves the overall objective function and it converges to a steady state solution. Furthermore, we propose a problem relaxation that reduces the computational burden associated with \cite{ILMPC} and satisfies real-time requirements.

As mentioned, we study also the case of model parameters uncertainty and their iterative estimation. In general, the MPC design under the presence of model uncertainty may be challenging. In \cite{c2} the author proposes to use two different models, a nominal one used to check the constraint satisfaction and a learnt one to improve performance. However, for the racing problem it may be difficult to define a priori a model that guaranties the feasibility. A different approach is proposed in \cite{c13}, where MPC algorithm uses a simple a priori vehicle model and a learned disturbance model. The authors proved the effectiveness of this control strategy for a path following problem. In the racing problem, we would like to have a model of vehicle when operating close to its handling capability. Unfortunately, modelling the vehicle lateral dynamics is challenging. Also when the linear model assumption holds, the 
tire stiffness are speed and environment dependent. In \cite{SetMembership} the lateral dynamics are estimated using a LPV system and a set membership algorithm has been successfully implemented. This procedure might be computationally demanding for on-line model learning. In this paper we propose an identification approach which exploits information from the previous iterations. At each time step $t$ of the MPC algorithm, system dynamics are estimated using only input-output data from previous iterations close to the current system state $x(t)$.

This paper is organized as follows: in Section II we introduce the vehicle model and we formalize the minimum time problem. The iterative formulation is illustrated in Section III, where we introduce the quantities necessary to implement the LMPC. In Section IV, the system identification procedure is described. The LMPC and its relaxation are illustrated in Section V. Finally, in Section VI we test the proposed control logic on an a section of a race track. Section VII provides final remarks.

\section{PROBLEM DEFINITION}
\subsection{Vehicle Model}
The vehicle dynamics is described by,
\begin{equation}\label{eq:VehicleModel}
\begin{bmatrix}
x_{t+1}\\
y_{t+1}
\end{bmatrix} = f(x_t, u_t)
\end{equation}
where $f(x_t, u_t)$ is the vehicle dynamic state update equation. The vectors $x_t$ and $u_t$ in (\ref{eq:VehicleModel}) collect the states and inputs of the vehicle at time $t$,
\begin{subequations}\label{eq:StateInputVector}
\begin{align}
x_t &= [v_{x_t} ~ v_{y_t} ~ \dot{\psi}_t ~ e_{\psi_t} ~ e_{y_t} ~ s_t ]\\
u_t &= [a_{t} ~ \delta_t ],
\end{align}
\end{subequations}
where $s_t$ represents the distance travelled along the center-line of the road, $e_{y_t}$ and $e_{\psi_t}$ the lateral distance and heading angle error between the vehicle and the path. $v_{x_t}$, $v_{y_t}$ and $\dot{\psi}_{t}$ are the vehicle longitudinal velocity, lateral velocity and yaw rate, respectively. The inputs are the longitudinal acceleration $a_t$ and the steering angle $\delta_t$. For more details on the curvilinear abscissa reference frame we refer to \cite{micaelli}.

\begin{figure}[h!]
\centering
\includegraphics[width=0.9\columnwidth]{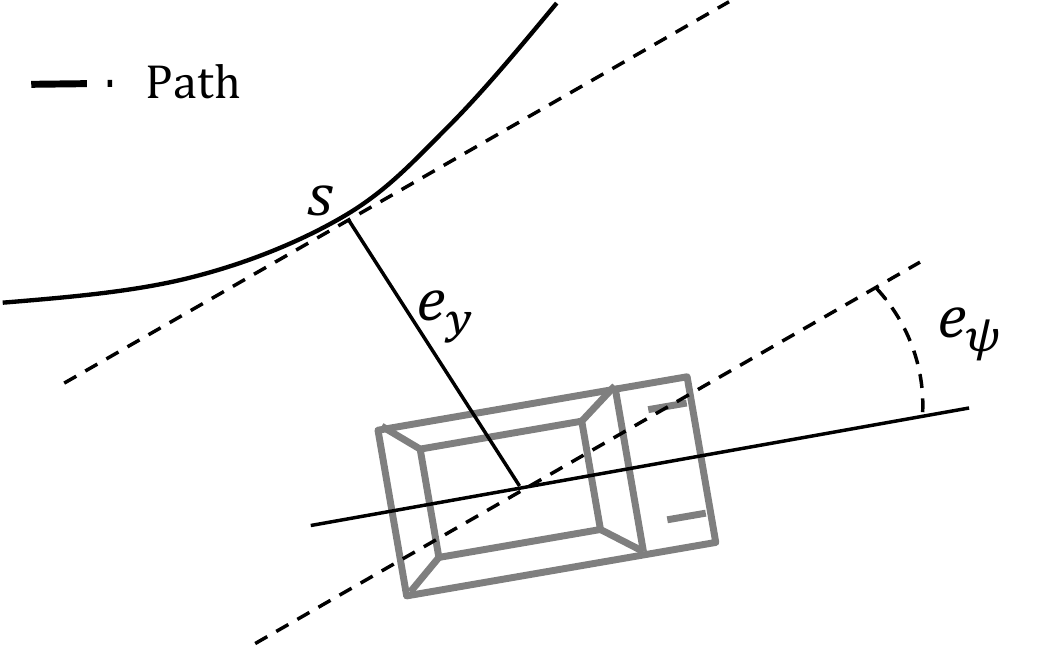}
\caption{Representation of the vehicle in the curvilinear abscissa reference frame}
\label{Fig:DynamicBicycleModel}
\end{figure}

\subsection{Minimum Time Problem}
The goal of the controller is to minimize the time to cross the finish line at $s_{target}$. More formally, the goal of the controller is to solve the following constrained infinite horizon optimal control problem

\begin{subequations}\label{eq:InfOCP}
\begin{align}
J_{0\rightarrow \infty}^*(x_S)&=\min_{u_0,u_1,\ldots} \sum\limits_{k=0}^{\infty} h(x_k,u_k)\label{eq:InfOCP1}\\
\textrm{s.t. }
&x_{k+1}=f(x_k,u_k),~\forall k\geq 0 \label{eq:InfOCP2}\\
&x_0=x_S,\label{eq:InfOCP3}\\
&x_k \in \mathcal{X},~u_k \in \mathcal{U},~\forall k\geq 0\label{eq:InfOCP4}
\end{align}
\end{subequations}
where equations (\ref{eq:InfOCP2}) and (\ref{eq:InfOCP3}) represent the vehicle dynamics and the initial condition, and (\ref{eq:InfOCP4}) are the state and input constraints. Note that In our problem the state constraint $\mathcal{X}$ is a convex set representing the road boundaries. The stage cost, $h(\cdot,\cdot)$, in equation (\ref{eq:InfOCP1}) is defined as
\begin{equation}\label{eq:CostInfOCP}
\begin{aligned}
h( x_k, u_k) = \begin{cases} 1 & \mbox{if } x_k \notin \mathcal{L} \\
0 & \mbox{if }  x_k \in \mathcal{L} \end{cases},
\end{aligned}
\end{equation}
where $\mathcal{L}$ is the set of points beyond the finish line at $s_{target}$,
\begin{equation}\label{eq:Def_x_f}
\begin{aligned}
\mathcal{L} = \Big\{x \in \rr^6 :~ x e_6^T = s > s_{target} \Big\}
\end{aligned}
\end{equation}
where $e_6$ is the the $6$-th standard basis in $\rr^6$.
\begin{assumption}
We assume that after the vehicle has crossed the finish line there exists a controller that can keep the vehicle on the road after the finish line. Namely, we assume that $\mathcal{L}$
is a control invariant set, thus
\begin{equation}\label{eq:Def_x_f1}
\begin{aligned}
\forall x_k \in \mathcal{L}, \exists u_k \in \mathcal{U} : x_{k+1}=f(x_k, u_k) \in \mathcal{L}.
\end{aligned}
\end{equation}
\end{assumption}

\section{LMPC PRELIMINARIES}
The constrained infinite horizon optimal control problem in (\ref{eq:InfOCP}) is difficult to solve, especially in real time. Therefore, we implemented the Learning MPC for iterative tasks presented in \cite{ILMPC}, where at each iteration the controller computes a trajectory that brings the system from the staring point $x_S$ to the invariant set $\mathcal{L}$. The data from each iteration are used to improve the performance of the controller until the controller converges to a local optimal solution, we refer to \cite{ILMPC} for further details. In the following we introduce the  notation that will be used to implement the Learning MPC for the racing application.\\

\subsection{Sampled Safe Set}
At the $j$-th iteration the vectors
\begin{subequations}\label{eq:sequence}
\begin{align}
{\bf{u}}^j ~ = ~ [u_0^j,~u_1^j,~...,~u_t^j,~...], \label{eq:sequenceU} \\
{\bf{x}}^j ~ = ~ [x_0^j,~x_1^j,~...,~x_t^j,~...], \label{eq:sequenceX}
\end{align}
\end{subequations}
collect the inputs applied to vehicle model~(\ref{eq:VehicleModel}) and the corresponding state, where $x_t^j$ and $u_t^j$ 
denote the vehicle state and the control input at time $t$ of the $j$-th iteration,
\begin{subequations}\label{eq:sequencetimet}
\begin{align}
x_t^j &= [v_{x_t}^j ~ v_{y_t}^j ~ \dot{\psi}_t^j ~ e_{\psi_t}^j ~ e_{y_t}^j ~ s_t^j ]\\
u_t^j &= [a_{t}^j ~ \delta_t^j ].
\end{align}
\end{subequations}
\begin{remark}
In (\ref{eq:sequenceX}) we have that $x_0^j = x_S,~\forall j\geq 0$, as the vehicle starts from the same initial point $x_S$ at each $j$-th iteration.
\end{remark}
We define the the \emph{sampled Safe Set} $\mathcal{SS}^j$ at iteration $j$ as
\begin{equation}\label{eq:SS}
\begin{aligned}
\mathcal{SS}^j = \textrm{}\left\{\bigcup_{i \in M^j} \bigcup_{t=0}^{\infty} x_t^i \right\}.
\end{aligned}
\end{equation}
$\mathcal{SS}^j$ is the collection of all state trajectories at iteration $i$ for $i\in M^j$. $M^j$ in equation (\ref{eq:SS}) is the set of indexes $k$ associated with successful iterations $k$ for $k\leq j$, defined as:
\begin{equation}\label{eq:M}
\begin{aligned}
M^j = \textrm{} \Big\{ k \in [0,j] : \lim_{t \to \infty} x_t^k = \mathcal{L} \Big\}.
\end{aligned}
\end{equation}

\subsection{Iteration Cost}
At time $t$ of the $j$-th iteration the cost-to-go associated with the closed loop trajectory (\ref{eq:sequenceX}) and input sequence (\ref{eq:sequenceU}) is defined as

\begin{equation}\label{eq:Functional}
\begin{aligned}
J_{t\rightarrow \infty}^j(x_t^j) = ~ \sum\limits_{k=t}^{\infty} h(x_k^j,u_k^j),
\end{aligned}
\end{equation}
where $h(\cdot,\cdot)$ is the stage cost of the problem (\ref{eq:InfOCP}).
We define the $j$-th iteration cost as the cost (\ref{eq:Functional}) of the $j$-th trajectory at time $t=0$,
\begin{equation}\label{eq:Performance}
\begin{aligned}
J_{0\rightarrow \infty}^j(x_0^j) = ~ \sum\limits_{k=0}^{\infty} h(x_k^j,u_k^j).
\end{aligned}
\end{equation}
$J_{0\rightarrow \infty}^j(x_0^j)$ quantifies the controller performance at each $j$-th iteration. \\

\begin{remark}
In equations (\ref{eq:Functional})-(\ref{eq:Performance}), $x_k^j$ and $u_k^j$ are the realized state and input at the $j$-th iteration, as defined in (\ref{eq:sequence}).\\
\end{remark}

\begin{remark}
At each $j$-th iteration the system is initialized at the same starting point $x_0^j = x_S$; thus we have $J_{0\rightarrow \infty}^j(x_0^j) = J_{0\rightarrow \infty}^j(x_S)$. \\
\end{remark}

Finally, we define the function $Q^j(\cdot)$, defined over the sampled safe set $\mathcal{SS}^j$ as:
\begin{equation}\label{eq:Qfunction}
\begin{aligned}
Q^j(x) = \begin{cases} \min\limits_{ (i,t) \in F^j(x)} J^i_{t\rightarrow \infty}(x), & \mbox{if } x \in \mathcal{SS}^j \\
~~~+\infty, & \mbox{if } x \notin \mathcal{SS}^j \end{cases},
\end{aligned}
\end{equation}
where $F^j(\cdot)$ in (\ref{eq:Qfunction}) is defined as
\begin{equation}\label{eq:Qfunction1}
\begin{aligned}
F^j(x) = \textrm{} \Big\{ (i,t) : i \in [0,j],~t\geq 0 ~&\textrm{with}~ x = x_t^i, \\
&~\textrm{for}~x_t^i \in \mathcal{SS}^j \Big\}.
\end{aligned}
\end{equation}
\begin{remark}
The function $Q^j(\cdot)$ in (\ref{eq:Qfunction}) assigns to every point in the sampled safe set, $\mathcal{SS}^j$, the minimum cost-to-go along the trajectories in $\mathcal{SS}^j$.
\end{remark}

\addtolength{\textheight}{-3cm}   

\section{SYSTEM IDENTIFICATION}
In this section the system identification technique is described. At the $j$-th iteration, the trajectories in the sampled safe set, $\mathcal{SS}^{j-1}$, and the data from the current iteration are used to estimate the system dynamics. In particular we assume that the system dynamic update at time $t$ of the $j$-th iteration is given by
\begin{equation}\label{eq:VehicleModelUpdate}
x^j_{t+1} = f_t^j(x_t^j, u_t^j) = \bar{g}(x_t^j, u_t^j) + g_t^j(x_t^j, u_t^j)
\end{equation}
where $\bar{g}(x_t^j, u_t^j) \in \rr^n$ is known function that represents the known dynamic of the system, and at time $t$ of the $j$-th iteration the function
\begin{equation}\label{eq:ModelRegressor}
g_t^j(x_t^j, u_t^j) = [g_{1,t}^j(\cdot), \cdots, g_{n,t}^j(\cdot)]^T\in \rr^n
\end{equation} is identified using a least mean square technique. The entries of $g_t^j(\cdot)$ are defined as 
\begin{equation}\label{eq:VehicleModelUpdateFunctions}
g_{i,t}^j(\cdot) = \gamma_{i,t}^j   \theta_{i,t}^j, ~i \in \{1, \cdots, n \}
\end{equation}
where $\gamma_{i,t}^j, ~i \in \{1, \cdots, n \}$ is the feature vector that may change for different states and $\theta_{i,t}^j,~\forall i \in  \{1, \cdots, n \}$ is the parameter vector which is estimated on-line. \\

\begin{remark}
We refer to the Appendix I for further details on how we implemented (\ref{eq:VehicleModelUpdate})-(\ref{eq:VehicleModelUpdateFunctions}) for the racing problem.\\
\end{remark}

Furthermore, in order to estimate the parameter vector $ \theta_{i,t}^j \in \rr^{p_i},~\forall i \in  \{1, \cdots, n \}$,
we implemented the following least mean square technique,
\begin{equation}\label{eq:LinearRegressorOpt}
\theta_{i,t}^j = \argminF_{\bar{\theta}} ||\phi_{i,t}^j \theta-y_{i,t}^j||_2, ~\forall i \in  \{1, \cdots, n \},
\end{equation}
where the vector $y_{t,i}^j$ selects a subset of the collected data using a distance-based criterion.
In particular, at time $t$ of the $j$-th iteration we define the time index $l^{t,j}_{k}$ with $k \leq j$, for which $x_{l^{t,j}_{k}}^{k}$ 
is the closest point to the current system state $x_t^j$,
\begin{equation}\label{eq:TimeIndex}
l^{t,j}_{k} = \argminF_{\tilde{l}} ||x_t^j - x_{\tilde{l}}^{k}||_2.
\end{equation}
Afterwards, the measurement vector, $y_{t,i}^j$, in (\ref{eq:LinearRegressorOpt}) is constructed using the data collected in the last $\bar{n}$ steps, and, for  the last $n_j\text{+1}$ trajectories in $\mathcal{SS}^{j-1}$, the data collected $\bar{n}$ steps before and $n$ steps after $l_k^{t,j}$,
\begin{equation} \label{eq:Measurements}
y_{i,t}^j = \begin{bmatrix}
\Big( x_{l_j^{t,j}}^j - \bar{g}(x_{l_j^{t,j}\text{-}1}, u_{l_j^{t,j}\text{-}1})\Big) e_i^T\\
\vdots \\
\Big( x_{l_j^{t,j}{\text-}\bar{n}}^j -  \bar{g}(x_{l_j^{t,j}{\text-}\bar{n}\text{-}1}^j, u_{l_j^{t,j}{\text-}\bar{n}\text{-}1}^j)\Big) e_i^T\\
~ \\\Big( x_{l_{j{\text-}1}^{t,j}{\text+}{n}}^{j{\text-}1} -\bar{g}(x_{l_{j{\text-}1}^{t,j}{\text+}{n}\text{-}1}^{j{\text-}1}, u_{l_{j{\text-}1}^{t,j}{\text+}{n}\text{-}1}^{j{\text-}1} ) \Big) e_i^T\\
\vdots \\
\Big( x_{l_{j{\text-}1}^{t,j}{\text-}\bar{n}}^{j{\text-}1} - \bar{g}(x_{l_{j{\text-}1}^{t,j}{\text-}\bar{n}\text{-}1}^{j{\text-}1}, u_{l_{j{\text-}1}^{t,j}{\text-}\bar{n}\text{-}1}^{j{\text-}1}) \Big) e_i^T\\
\vdots \\
\Big( x_{l_{j{\text-}n_j}^{t,j}{\text+}{n}}^{j{\text-}n_j} - \bar{g}(x_{l_{j{\text-}n_j}^{t,j}{\text+}{n}\text{-}1}^{j{\text-}n_j}, u_{l_{j{\text-}n_j}^{t,j}{\text+}{n}\text{-}1}^{j{\text-}n_j})\Big) e_i^T\\
\vdots \\
\Big( x_{l_{j{\text-}n_j}^{t,j}{\text-}\bar{n}}^{j{\text-}n_j} e_i^T- \bar{g}(x_{l_{j{\text-}n_j}^{t,j}{\text-}\bar{n}\text{-}1}^{j{\text-}n_j}, u_{l_{j{\text-}n_j}^{t,j}{\text-}\bar{n}\text{-}1}^{j{\text-}n_j})\Big) e_i^T\\
\end{bmatrix} \in \rr^{\bar{N}},
\end{equation}
$i \in \{1, \cdots, n \}$, where $e_i$ is the standard basis in $\rr^n$, and  $\bar{N} = \bar{n}\text{+}1 \text{+} n_j (\bar{n}\text{+}n\text{+}1)$ is the number of points used for the system identification. \\
\begin{remark}
$\bar{n}, n_j, n$ are tuning variables which are chosen by the designer.
\end{remark}
In (\ref{eq:LinearRegressorOpt}), the regressor matrix $\phi_{t,i}^j, \forall i \in [v_{x}, v_y, \dot{\psi}]$ is computed using the feature vectors,
and the data collected in $\mathcal{SS}^{j-1}$ and current iterations,
\begin{equation} \label{eq:RegressorMartix}
\phi_{i,t}^j = \begin{bmatrix}
\gamma_{i_{l_j^{t,j}\text{-}1}}^j \\
\vdots \\
\gamma_{i_{l_j^{t,j}{\text-}\bar{n}\text{-}1}}^j \\
~ \\
\gamma_{i_{l_{j{\text-}1}^{t,j}{\text+}{n}\text{-}1}}^{j\text{-}1} \\
\vdots \\
\gamma_{i_{l_{j{\text-}1}^{t,j}{\text-}\bar{n}\text{-}1}}^{j\text{-}1} \\
\vdots \\
\gamma_{i_{l_{j{\text-}n_j}^{t,j}{\text+}{n}\text{-}1}}^{j\text{-}n_j} \\
\vdots \\
\gamma_{i_{j{\text-}n_j}^{t,j}{\text-}\bar{n}\text{-}1}^{j\text{-}n_j} \\
\end{bmatrix} \in \rr^{\bar{N} \times p_i}, ~ i \in \{1, \cdots, n \}
\end{equation}

\section{LMPC CONTROL DESIGN}
\subsection{LMPC Formulation}

The LMPC tries to compute a solution to the infinite time optimal control problem (\ref{eq:InfOCP}) by solving at time $t$ of iteration $j$ the finite time constrained optimal control problem

\begin{subequations}\label{eq:Constraints}
\begin{align}
&J_{t\rightarrow t+N}^{\scalebox{0.4}{LMPC},j}(x_t^j)=\min_{u_{t|t},\ldots,u_{t+N-1|t}} \bigg[  \sum_{k=t}^{t+N-1}  h(x_{k|t},u_{k|t}) +\notag\\
&~~~~~~~~~~~~~~~~~~~~~~~~~~~~~~~~~~~~~~~~~~~~~~~+ Q^{j-1}(x_{t+N|t}) \bigg]\\
&\textrm{s.t. }\notag \\
&~~~x_{k+1|t}=f_t^j(x_{k|t},u_{k|t}),~\forall k \in [t, \cdots, t+N-1] \label{eq:Constraints1}\\
&~~~x_{t|t}=x_t^j,\label{eq:Constraints2}\\
&~~~x_{k|t} \in \mathcal{X}, ~ u_k \in \mathcal{U},~\forall k \in [t, \cdots, t+N-1] \label{eq:Constraints4}\\
&~~~x_{t+N|t} \in ~\mathcal{SS}^{j-1},\label{eq:Constraints5}
\end{align}
\end{subequations}
where (\ref{eq:Constraints1}) and (\ref{eq:Constraints2}) represent the system dynamics in (\ref{eq:VehicleModelUpdate}) and initial condition, respectively. The state and input constraints are given by (\ref{eq:Constraints4}). Finally (\ref{eq:Constraints5}) forces the terminal state into the set $\mathcal{SS}^{j-1}$ defined in equation (\ref{eq:SS}).\\
Let
\begin{equation}\label{eq:OptimalSolutionMPC}
\begin{aligned}
{\bf{u}}^{*,j}_{t:t+N|t}  &= [u_{t|t}^{*,j}, \cdots, u_{t+N-1|t}^{*,j}]\\
{\bf{x}}^{*,j}_{t:t+N|t} &= [x_{t|t}^{*,j}, \cdots, x_{t+N|t}^{*,j}]
\end{aligned}
\end{equation}
be the optimal solution of (\ref{eq:Constraints}) at time $t$ of the $j$-th iteration and $J_{0\rightarrow N}^{\scalebox{0.4}{LMPC},j}(x_t^j)$ the corresponding optimal cost. Then, at time $t$ of the iteration $j$, the first element of ${\bf{u}}^{*,j}_{t:t+N|t}$ is applied to the system (\ref{eq:VehicleModel})
\begin{equation}\label{eq:MPC}
\begin{aligned}
u_t^j = u_{t|t}^{*,j}.
\end{aligned}
\end{equation}
The finite time optimal control problem (\ref{eq:Constraints}) is repeated at time $t+1$, based on the new state $x_{t+1|t+1} = x_{t+1}^j$ (\ref{eq:Constraints2}), yielding a \textit{moving} or \textit{receding horizon} control strategy.\\

\subsection{LMPC Relaxation}
As the sampled safe set in (\ref{eq:SS}) is a set of discrete points, the optimal control problem in (\ref{eq:Constraints}) is a Nonlinear Mixed Integer Programming, therefore the LMPC (\ref{eq:Constraints}) and (\ref{eq:MPC}) may be computationally challenging to solve in real-time. Therefore, we introduce the time varying approximated safe set, $\mathcal{\tilde{SS}}^{ j-1}_t$, and the time varying approximated $Q^{j-1}(\cdot)$ function, $\tilde{Q}^{j-1}_t(\cdot)$. At time $t$ of the $j$-th iteration, we approximate $\mathcal{SS}^{j-1}$ with the time varying approximated sampled safe set,
\begin{equation}\label{eq:SSRelax}
\begin{aligned}
\mathcal{\tilde{SS}}^{ j-1}_t = \Big\{ x \in \rr^6, \lambda &\in [0, 1] : 
\\ & x \in \lambda \tilde{{\bf{x}}}^{ j-1}_t + (1-\lambda) \tilde{{\bf{x}}}^{ j-2}_t  \Big\} 
\end{aligned}
\end{equation}
where 	$\tilde{{\bf{x}}}^{ j-1}_t$ approximates locally the $j\text{-}1 \text{-}$th trajectory, ${\bf{x}}^{j-1}$, using a $5$-th order polynomial function,
\begin{equation}\label{eq:x^jContinousTime}
\begin{aligned}
\tilde{{\bf{x}}}^{j-1}_t =  \Big\{  x \in \rr^6 : ~\forall i& \in \{v_x, ~v_y,~ \dot{\psi},~ e_{\psi},~ e_y \}, \\
&~ i =~ [s^5~s^4~s^3~ s^2~s ~ 1]~\Gamma_{t,i}^{j-1} \Big\},
\end{aligned}
\end{equation}
being $\Gamma_{t,i}^{j-1}$ the solution to the following least mean square problem
\begin{equation}\label{eq:x^jContinousTimeRegression}
\begin{aligned}
\Gamma_{t,i}^{j-1} = \small{  \argminF_\Gamma ||}
&\begin{bmatrix}
i_{l_{j-1}^{t,j}+4N}^{j-1} \\
\vdots \\
i_{l_{j-1}^{t,j}}^{j-1} \\
\end{bmatrix} + \\
&~~~~~-\begin{bmatrix}
s^5_{l_{j{\text-}1}^{t,j}\text{+}4\text{N}} & \cdots & s_{l_{j{\text-}1}^{t,j}\text{+}4\text{N}} & 1 \\
\vdots        &~        ~&\vdots       &\\
s^5_{l_{j{\text-}1}^{t,j}} &\cdots & s_{l_{j{\text-}1}^{t,j}} & 1\\
\end{bmatrix} \Gamma \small{||_2,}
\end{aligned} 
\end{equation}
where $l_{j{\text-}1}^{t,j}$ is the time index defined in equation (\ref{eq:TimeIndex}).
\begin{remark}
Note that $\tilde{{\bf{x}}}^{ j-2}_t$ is defined replacing $j\text{-}1$ with $j\text{-}2$ in the above definition.
\end{remark}

Furthermore, we introduce the time varying function $C^{j-1}_t(\cdot)$, which at time $t$ of the $j$-th iteration approximates the cost to go along the $j\text{-}1 \text{-}$th trajectory,
\begin{equation}\label{eq:Cfunction}
\begin{aligned}
C^{j-1}_t(x) = \begin{cases} [s^5~s^4~s^3~ s^2~s ~ 1] \Delta^{j-1}_t, & \mbox{if } x \in \tilde{{\bf{x}}}^{j-1}_t\\
~~~+\infty, & \mbox{if } x \notin \tilde{{\bf{x}}}^{j-1}_t \end{cases},
\end{aligned}
\end{equation}
where $\Delta_{t,i}^{j-1}$ is the solution to the following least mean square problem
\begin{equation}\label{eq:CfunctionRegression}
\begin{aligned}
\Delta^{ j-1}_t = \small{  \argminF_\Delta ||} &
\begin{bmatrix}
J_{t\rightarrow \infty}^{j-1}(x_{l_{j{\text-}1}^{t,j}\text{+}4\text{N}}^{j\text{-}1}) \\
\vdots \\
J_{t\rightarrow \infty}^{j-1}(x_{l_{j{\text-}1}^{t,j}}^{j\text{-}1}) & \\
\end{bmatrix} + \\
&~~~~~ -\begin{bmatrix}
s^5_{l_{j{\text-}1}^{t,j}\text{+}4\text{N}} & \cdots & s_{l_{j{\text-}1}^{t,j}\text{+}4\text{N}} & 1 \\
\vdots        &~        ~&\vdots       &\\
s^5_{l_{j{\text-}1}^{t,j}} &\cdots & s_{l_{j{\text-}1}^{t,j}} & 1\\
\end{bmatrix} \Delta \small{||_2.}
\end{aligned} 
\end{equation}
\begin{remark}
Note that $C^{j-2}_t(\cdot)$ is defined replacing $j\text{-}1$ with $j\text{-}2$ in the above definition.
\end{remark}
The $C^{j-1}_t(\cdot)$ function in (\ref{eq:Cfunction}) is used to define the continuous time varying approximation of ${Q}^{j-1}(\cdot)$,
\begin{equation}\label{eq:QFunctionApproximation}
\begin{aligned}
\tilde{Q}^{j-1}_t(x, \lambda) = \begin{cases} \lambda C^{ j-1}_t(x) +& (1-\lambda)C^{ j-2}_t(x), \\
& ~~~~~~\mbox{if } (x, \lambda) \in \mathcal{\tilde{SS}}^{j-1}_t\\
~~~+\infty, 
&  ~~~~~~\mbox{if } (x, \lambda) \in \mathcal{\tilde{SS}}^{j-1}_t
\end{cases}.
\end{aligned}
\end{equation}

Finally, we reformulate the LMPC in (\ref{eq:Constraints}) and (\ref{eq:MPC}) using the time varying approximation of $\mathcal{SS}^{j-1}$ and of the $Q^{j-1}(\cdot)$ function, to have a computationally tractable problem. We define the following constrained finite time optimal control problem,
\begin{subequations}\label{eq:RelaxedLMPC}
\begin{align}
\tilde{J}_{t\rightarrow t+N}^{\scalebox{0.4}{LMPC},j}&(x_t^j)=\min_{\lambda, u_{t|t},\ldots,u_{t+N-1|t}} \bigg[  \sum_{k=t}^{t+N-1}  h(x_{k|t},u_{k|t}) +\notag\\
&~~~~~~~~~~~~~~~~~~~~~~~~~~~~~~~~~~+ \tilde{Q}^{j-1}_t(x_{t+N|t}, \lambda) \bigg]\\
\textrm{s.t. }&\notag \\
&x_{k+1|t}=f_t^j(x_{k|t},u_{k|t}),~\forall k \in [t, \cdots, t+N-1] \label{eq:RelaxedLMPC1}\\
&x_{t|t}=x_t^j,\label{eq:RelaxedLMPC2}\\
&x_{k|t} \in \mathcal{X}, ~ u_k \in \mathcal{U},~\forall k \in [t, \cdots, t+N-1] \label{eq:RelaxedLMPC3}\\
&(x_{t+N|t}, \lambda) \in ~\mathcal{\tilde{SS}}^{ j-1}_t. \label{eq:RelaxedLMPC4}
\end{align}
\end{subequations}

Let
\begin{equation}\label{eq:OptimalSolutionMPC1}
\begin{aligned}
{\bf{\tilde{u}}}^{*,j}_{t:t+N|t}  &= [\tilde{u}_{t|t}^{*,j}, \cdots, \tilde{u}_{t+N-1|t}^{*,j}]\\
{\bf{\tilde{x}}}^{*,j}_{t:t+N|t} &= [\tilde{x}_{t|t}^{*,j}, \cdots, \tilde{x}_{t+N|t}^{*,j}]
\end{aligned}
\end{equation}
be the optimal solution of (\ref{eq:RelaxedLMPC}) at time $t$ of the $j$-th iteration and $\tilde{J}_{0\rightarrow N}^{\scalebox{0.4}{LMPC},j}(x_t^j)$ the corresponding optimal cost. Then, at time $t$ of the iteration $j$, the first element of ${\bf{\tilde{u}}}^{*,j}_{t:t+N|t}$ is applied to the system (\ref{eq:VehicleModel})
\begin{equation}\label{eq:MPC1}
\begin{aligned}
u_t^j = \tilde{u}_{t|t}^{*,j}.
\end{aligned}
\end{equation}
The finite time optimal control problem (\ref{eq:RelaxedLMPC}) is repeated at time $t+1$, based on the new state $x_{t+1|t+1} = x_{t+1}^j$ (\ref{eq:RelaxedLMPC2}).\\

\section{SIMULATION RESULTS}
Simulations are performed in CarSim which uses a high-fidelity vehicle model to simulate the vehicle dynamics. The parameters of the CarSim vehicle and tire models were identified using data collected from our experimental passenger vehicle. The nonlinear optimization problem \eqref{eq:RelaxedLMPC} is solved using NPSOL \cite{NPSOL}. To verify the real-time feasibility of the LMPC strategy, we successfully ran the closed-loop simulation on a dSpace MicroAutobox II embedded computer ($900$MHz IBM PowerPC processor) at a sampling time of $100$ms. The LMPC is discretized at $100$ms and the horizon $N = 10$. The system identifiation algorithms is implemented using $n = \bar{n} = 50$ and $n_j=2$. Future work will focus on the implementation on our experimental vehicle.

In order to implement the LMPC (\ref{eq:RelaxedLMPC}) and (\ref{eq:MPC1}), we computed a feasible trajectory, ${\bf{x^0}}$, that drives the from $x_S$ to the invariant set $\mathcal{L}$ using a path following controller at a low velocity. For more details on the path following controller, we refer to \cite{Master}. The first feasible trajectory, ${\bf{x^0}}$, is used to construct $\mathcal{\tilde{SS}}^{0}_t$ and $\tilde{Q}^{0}_t$, needed to initialize the first iteration of the LMPC (\ref{eq:RelaxedLMPC}) and (\ref{eq:MPC1}). Parameters used in the controller are reported in Table I.

\begin{figure}[h!]
	\centering
	\includegraphics[width=0.9\columnwidth]{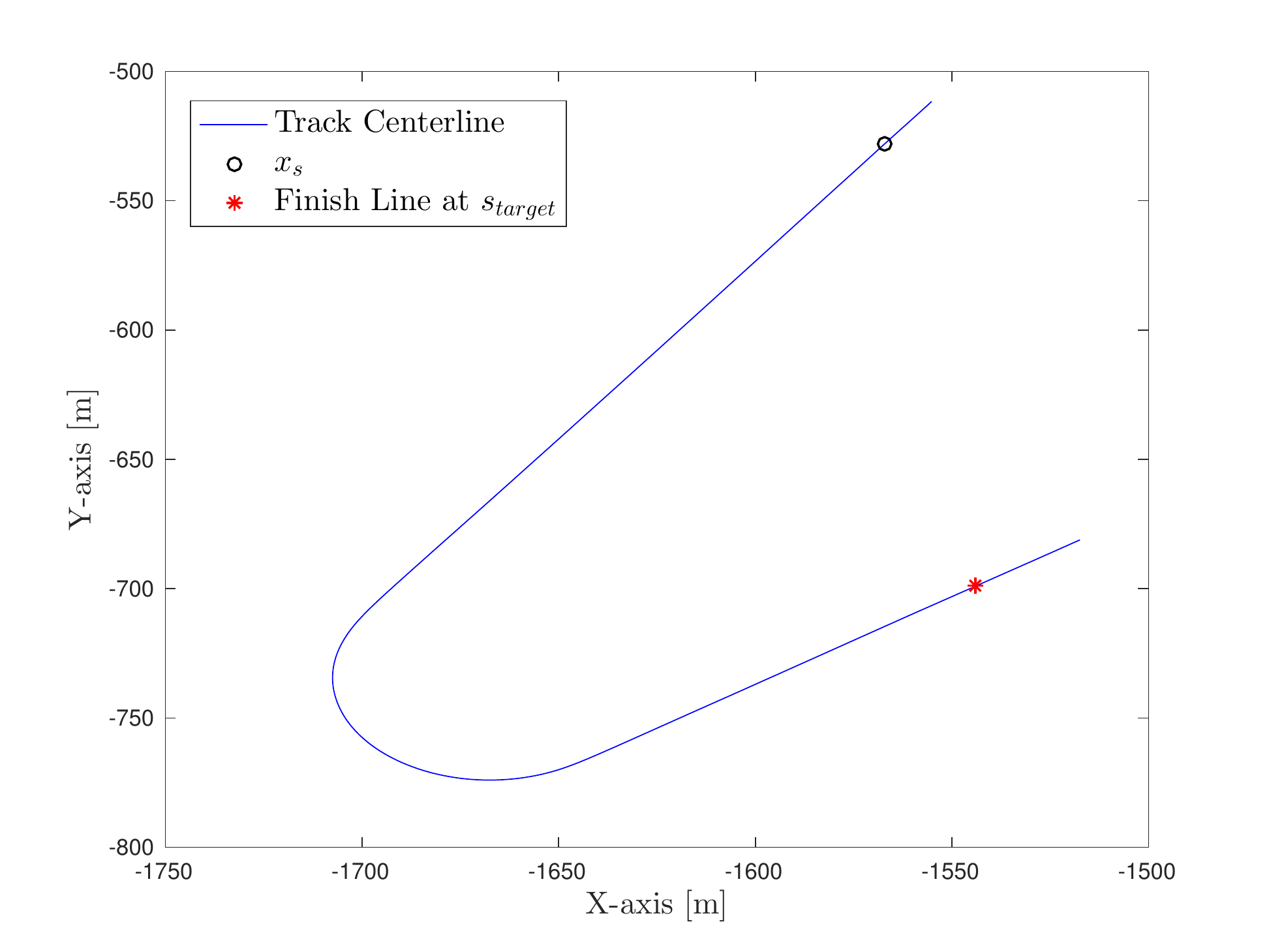}
	\caption{Path used for testing the proposed control logic.}
	\label{Res1:Path}
\end{figure}

To effectively illustrate the results, the controller is tested on a single corner of a race track. The track centerline, starting position and finish line are shown in Figure \ref{Res1:Path}. In particular, we have a straight in the first section, a curve in the second section and a straight again in the third section. We define the time $\tilde{t}_j$ at which the vehicle crosses the finish line at the $j$-th iteration as \begin{equation}
\tilde{t}_j = \min \{t \in \zz_{0+} : x^j_t \in \mathcal{L} \}.
\end{equation}
The track borders are described by the box constraint $|e_y| \leq 1.6 m$. Note that we assumed that the road constraint on the lateral distance $e_y$ takes into account the width of the vehicle. 

\begin{figure}[h!]
		\centering
		\includegraphics[width=0.9\columnwidth]{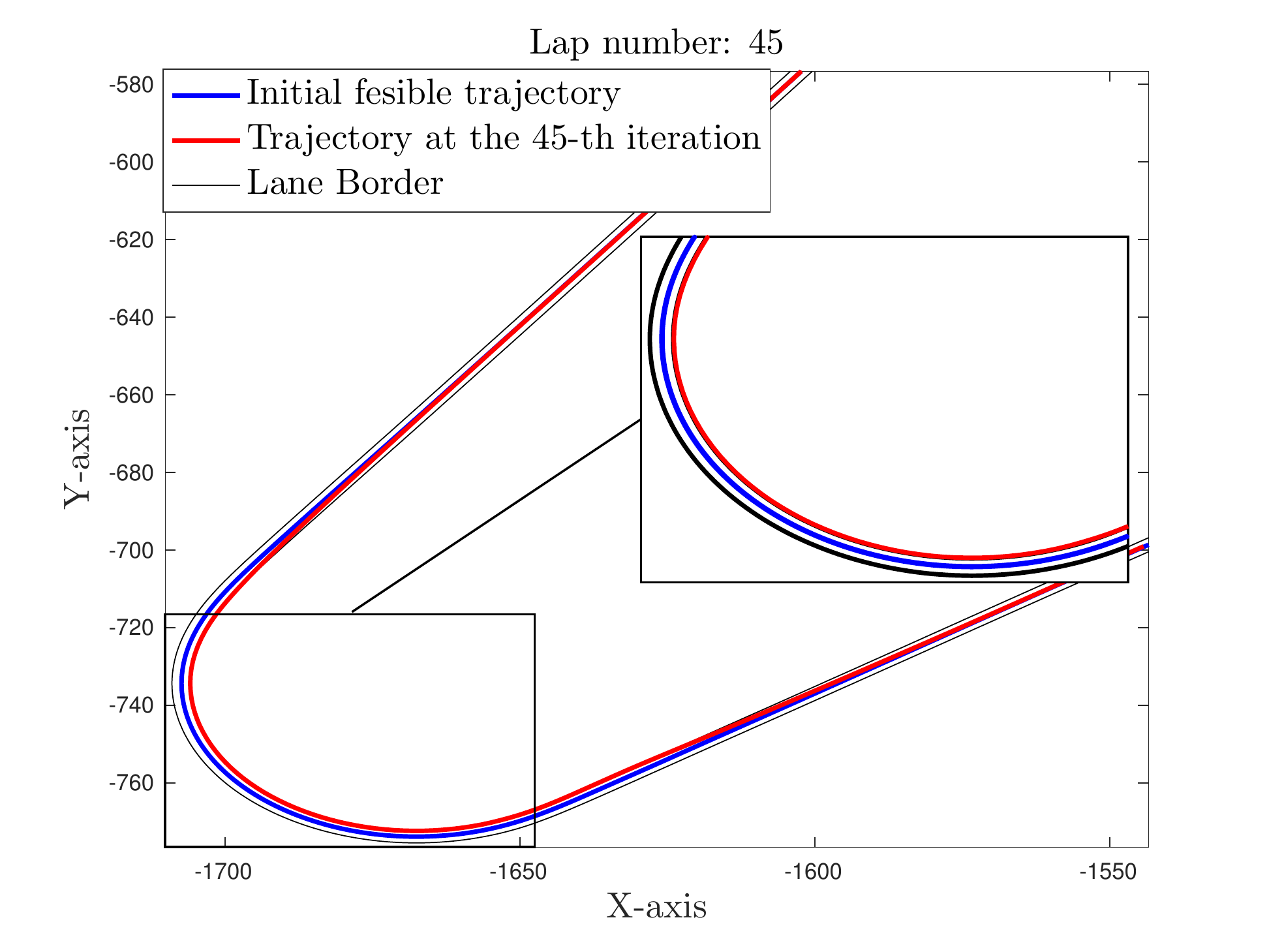}
		\caption{Steady state trajectory of the proposed control strategy on the $X-Y$ plane.}
		\label{Res1:SteadyStateTrajectory}
	\end{figure}

\begin{figure*}[h!]
		\includegraphics[width=1.0\textwidth]{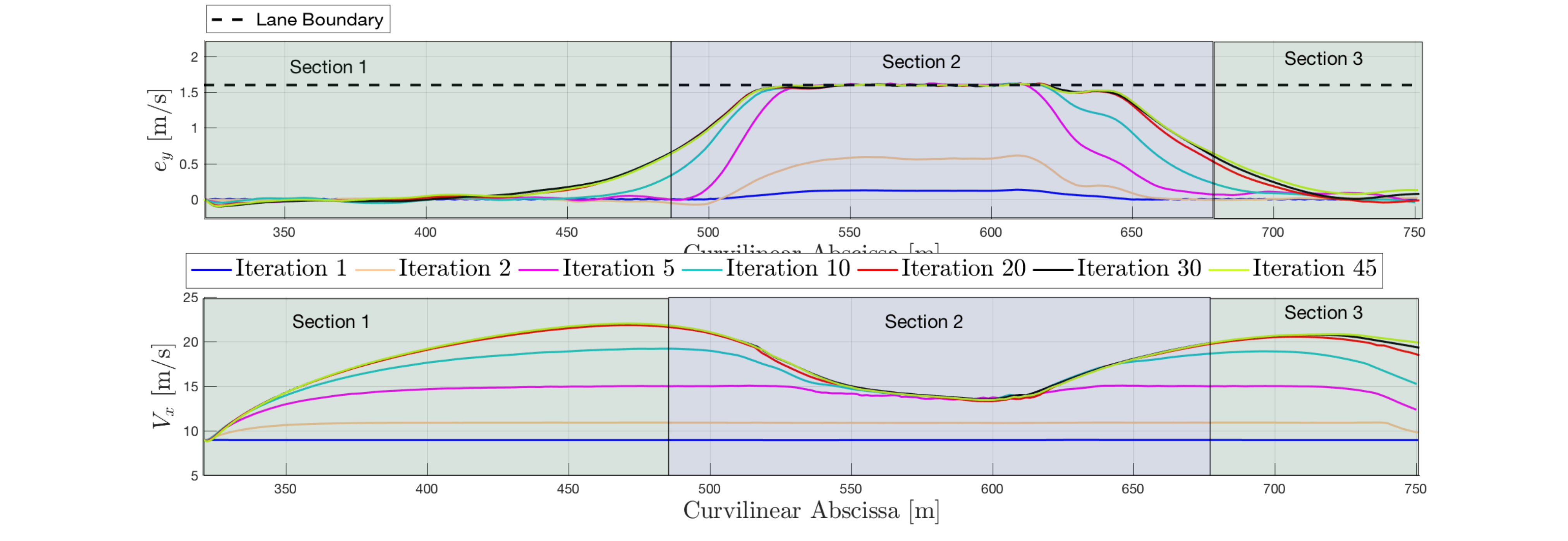}
		\caption{Evolution of the closed-loop trajectory over the iterations}
		\label{Res1:TrajectoryEvolution}
	\end{figure*}

The initial feasible trajectory and the locally optimal trajectory that the controller converges to are shown in Figure \ref{Res1:SteadyStateTrajectory}.
It is seen that the vehicle cuts the corner performing a trajectory with a constant radius of curvature.
In particular, we see that the controller saturates the road constraint in the second section of the path, which represents the curve (Fig. \ref{Res1:TrajectoryEvolution}). This behavior was shown to be the optimal solution for the minimum time problem racing problems \cite{MinTime}.

\begin{figure}[h!]
		\centering
		\includegraphics[width=0.9\columnwidth]{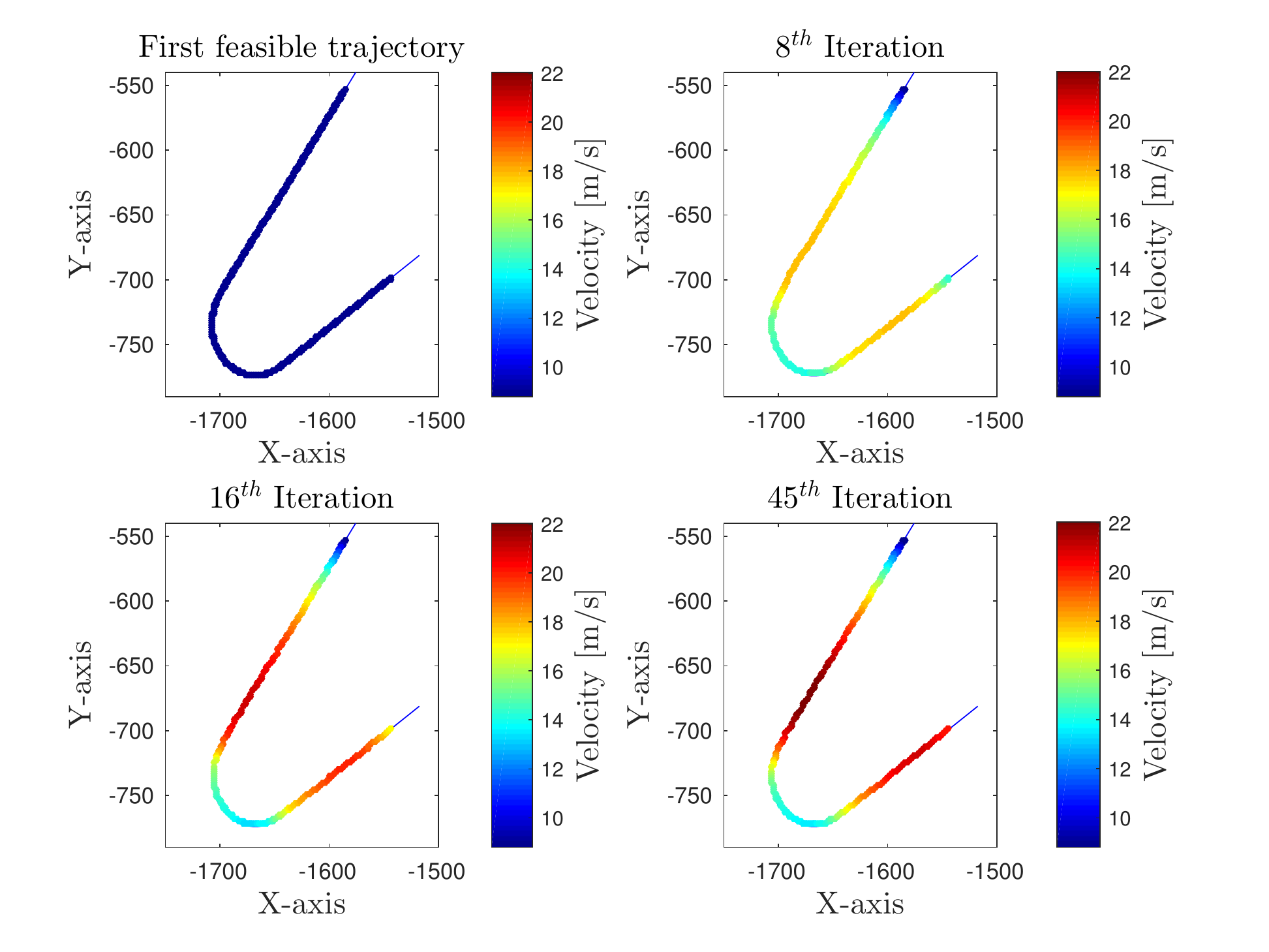}
		\caption{Evolution of the velocity profile over the iterations.}
		\label{Res1:LapVelocityProfile}
	\end{figure}

Figure \ref{Res1:LapVelocityProfile} depicts the velocity profile along the trajectory. The controller slows down just before entering corner, speeds up right after the midpoint of the curve. This behavior is consistent with racing driver performance \cite{theodosis2011generating} \cite{Rucco}.

\begin{figure}[h!]
		\centering
		\includegraphics[width=0.9\columnwidth]{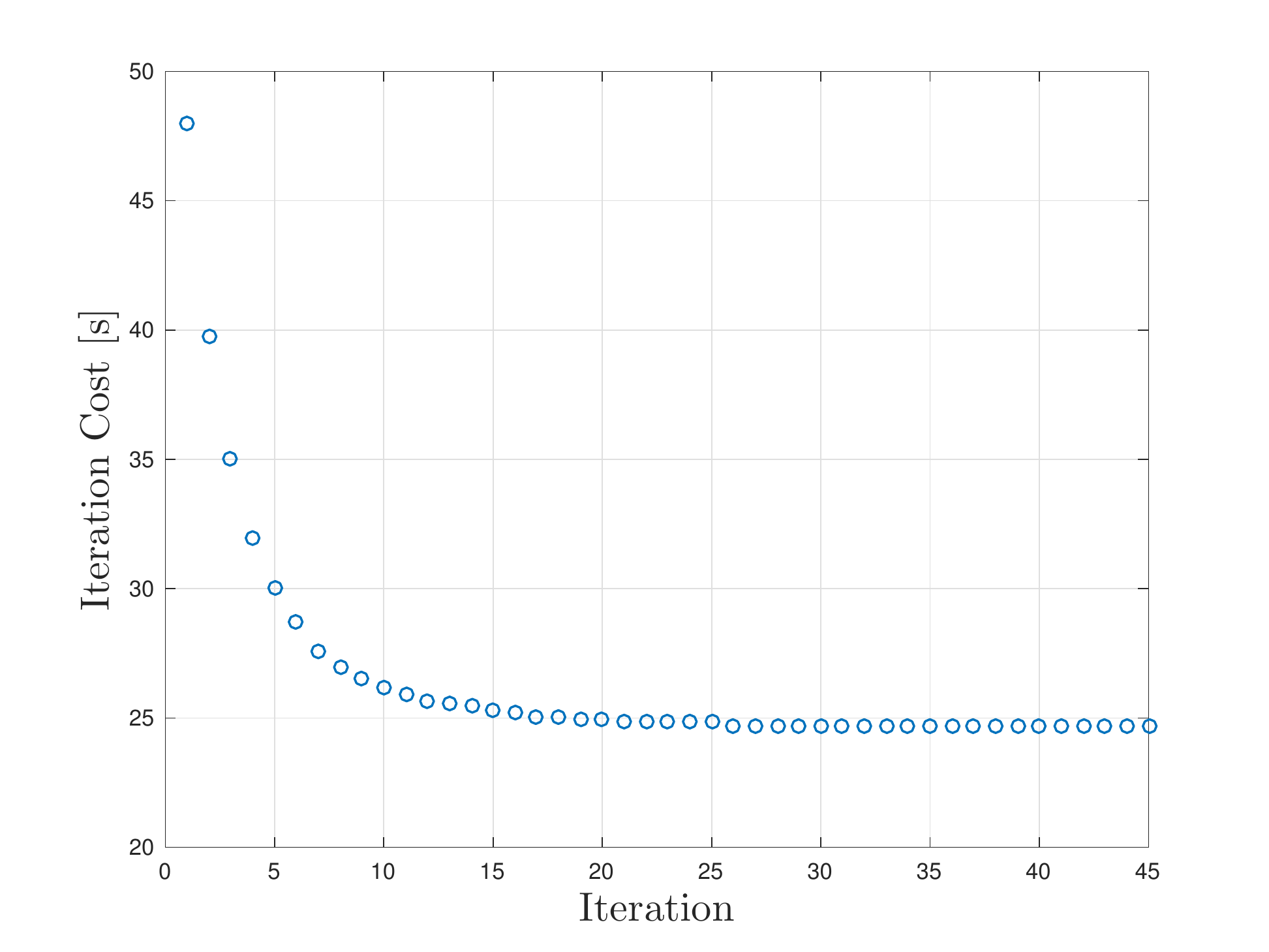}
		\caption{Evolution of the iteration cost over the iterations. We notice that the LMPC coupled with the proposed system identification technique decreases the travelling time at each iteration.}
		\label{Res1:IteartionCost}
	\end{figure}

The controller successfully decreases the traveling time from the starting point, $x_S$, to the finish line in Figure \ref{Res1:Path}. Furthermore, we see in Figure \ref{Res1:IteartionCost} that the travelling time is decreasing at each iteration until the LMPC (\ref{eq:RelaxedLMPC}) and (\ref{eq:MPC1})  reaches convergence.

Figure \ref{Res1:TrajectoryEvolution} shows the evolution of the vehicle longitudinal velocity and lateral distance from the center-line. The controller correctly understands the benefit of cutting the curve until it saturates the road constraints. Moreover, the controller brakes until the midpoint of the curve to prevent the vehicle to drift out of the track.

	\begin{figure}[h!]
		\centering
		\includegraphics[width=1.0\columnwidth]{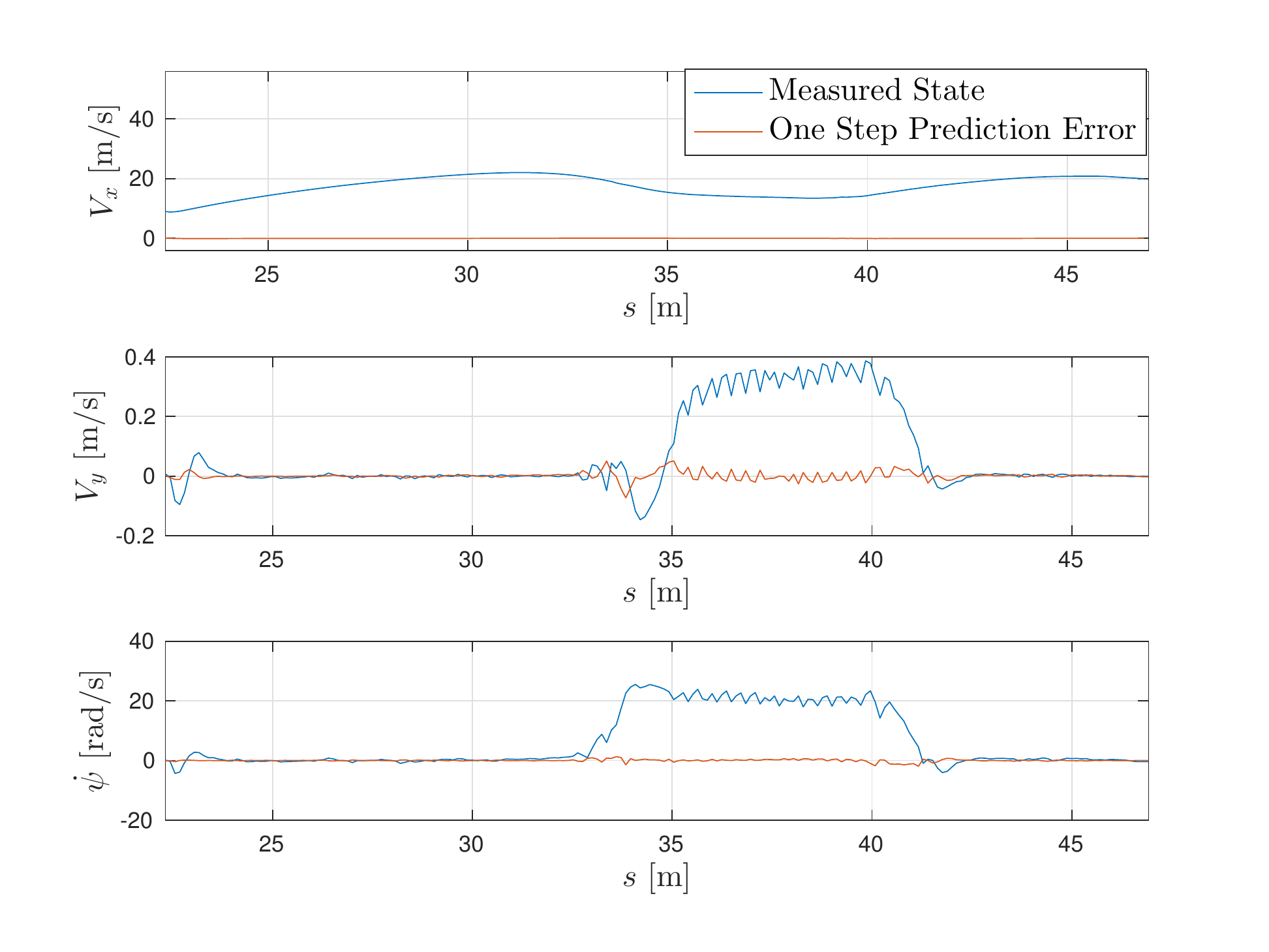}
		\caption{Comparision between the prediciton error and the closed-loop trajectory.}
		\label{Res1:OneStep}
	\end{figure}
We analyse the one step prediction errors at time $t$ of the $j$-th iteration,
\begin{equation}
\nabla i_t^j = i^{*,j}_{t+1|t} - i_{t+1}^j, ~ \forall i \in [v_x, v_y, \dot{\psi}]
\end{equation}
which quantifies the model mismatch between the real model (\ref{eq:VehicleModel}) and the learnt model (\ref{eq:VehicleModelUpdate}). Figure \ref{Res1:OneStep} reports the one step prediction error for the $45$-th iteration, compared with the actual system state. We notice that the proposed distance-based identification technique correctly identifies the system dynamics within an acceptable tolerance, in particular we have,
\begin{subequations} \notag
\begin{align}
\max_t |\nabla v_{x_t}^j| &= 0.0587 ~\small{m / s} \\
\max_t |\nabla v_{y_t}^j| &= 0.0511~ \small{m / s} \\
\max_t |\nabla \dot{\psi}_{t}^j| &= 0.0224 ~\small{rad / s}.
\end{align}
\end{subequations}

	\begin{figure}[h!]
		\centering
		\includegraphics[width=0.9\columnwidth]{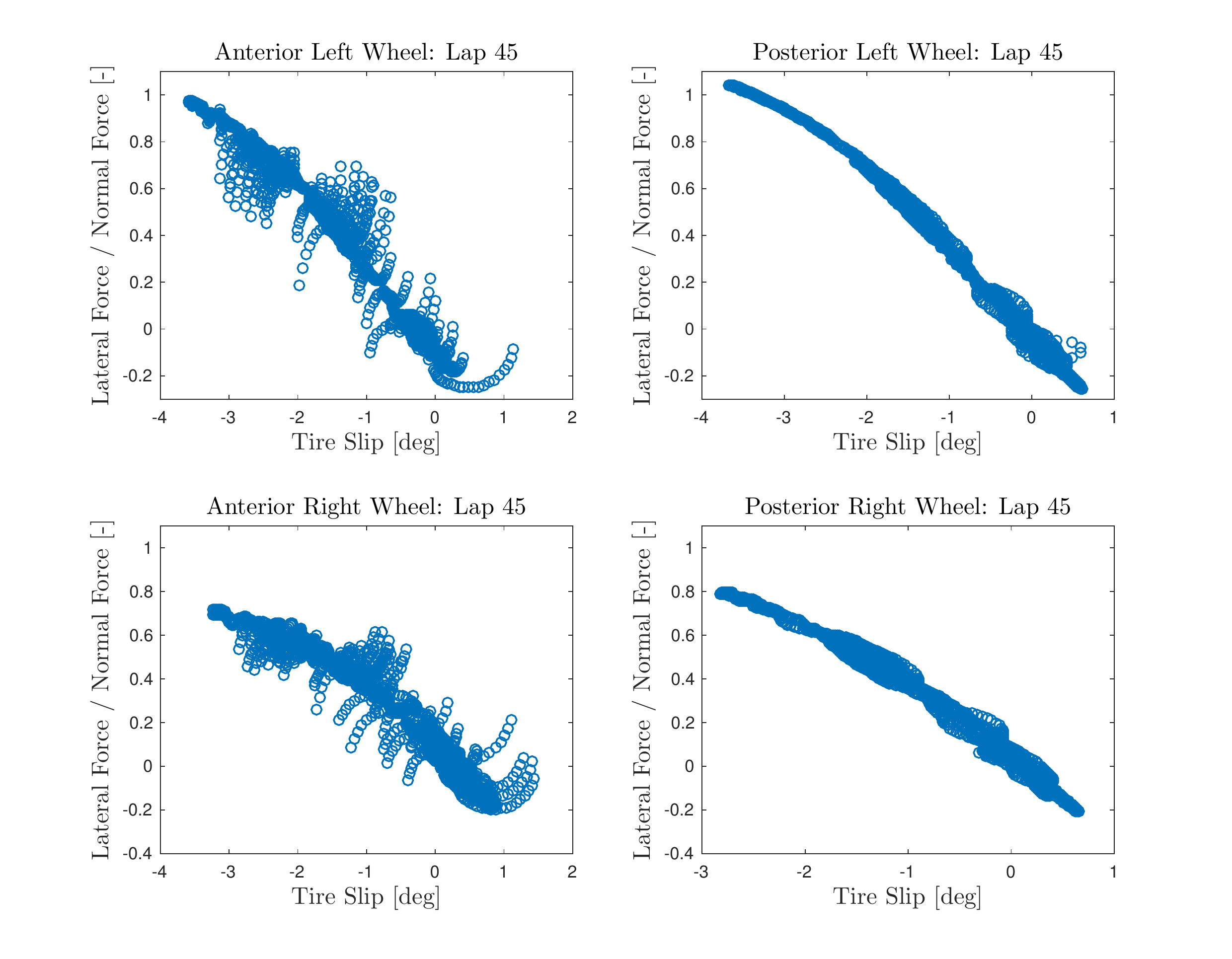}	\caption{Analysis of the tire forces. We notice that the left wheels are operating at their handling capability.}
		\label{Res1:TireForce}
	\end{figure}
Finally, Figure \ref{Res1:TireForce} shows that LMPC (\ref{eq:RelaxedLMPC}) and (\ref{eq:MPC1})  saturates the tire capabilities of the left front and rear tire. Therefore, we conform that the proposed control logic is able to identify the vehicle's performance limit and operate the vehicle at the limit of its handling capability.

\section{Conclusions}
In this paper, a learning nonlinear model predictive control for the racing problem that exploits information from the previous laps to improve the performance of the closed loop system over iterations is presented. A time varying approximation safe set and a terminal cost, learnt from previous iterations, are presented. Moreover, we coupled the LMPC with a distance-based identification method that allows the controller to operate the vehicle at the limit of its handling capability. We tested the proposed control logic in simulation with the high fidelity simulator CarSim and we showed that the controller is able to identify the vehicle dynamics and to operates the vehicle close to the limit of its handling capability.

\section{APPENDIX I}
This Section describes the model used for the identification techniques presented in Section IV. We assumed that the known dynamics in (\ref{eq:VehicleModelUpdate})  is given by the kinematic model in the error reference frame,
\begin{equation}
\renewcommand\arraystretch{2.0}
\begin{aligned}
 \bar{g}_t^j(x_i^j, u_i^j) =x_i^j +  \small {\begin{bmatrix}
	0 \\
	0\\
	0\\
	 \bigg[ \dot{\psi}_t - \frac{v_{x_t} cos(e_{\psi_t}) - 
		v_{y_t} sin(e_{\psi_t})}{1-e_{y_t} \kappa(s_t)} \kappa(s_t) \bigg] dt \\
	(v_{x_k} sin(e_{\psi_t}) + v_{y_t} cos(e_{e_{\psi_t}}))dt\\
	\frac{v_{x_k} cos(e_{\psi_t}) - v_{y_k} sin(e_{\psi_t})}{1-e_{y_k} \kappa(s_k)}dt
	\end{bmatrix}, }
\end{aligned}
\end{equation}
where $\kappa(s)$ is the angle of the tangent vector to the path at the curvilinear abscissa $s_t$. The states $e_{y_t}$ and $e_{\psi_t}$ represent the lateral distance and heading angle error between the vehicle and the path. $v_{x_t}$, $v_{y_t}$ and $\dot{\psi}_{t}$ are the vehicle longitudinal velocity, lateral velocity and yaw rate, respectively. \\
Furthermore, the linear regressor in (\ref{eq:ModelRegressor}) is used to identify the longitudinal and lateral dynamics and it is given by
\begin{equation}\label{eq:VehicleModelUpdateImplementation}
\renewcommand\arraystretch{2.0}
\begin{aligned}
g_t^j(x_i^j, u_i^j) =	\small{\begin{bmatrix}
	g_{t,1}^j(x_i^j, u_i^j)\\
	g_{t,2}^j (x_i^j, u_i^j)  \\
	g_{t,3}^j (x_i^j, u_i^j)\\
	0 \\
	0 \\
	0
	\end{bmatrix},} 
\end{aligned}
\end{equation}
where the feature vectors in (\ref{eq:VehicleModelUpdateFunctions})  are defined as
\begin{subequations}\label{eq:LinearRegressor_app}
\begin{align}
\gamma_{1,t}^j &= [v_{x_t}^j, v_{y_t}^j \dot{\psi}_t^j, a_t^j] ~~~~~~~~~~~~ \in \rr^{3} \\
\gamma_{2,t}^j &= \Big[\frac{v_{y_t}^j}{v_{x_t}^j} , \dot{\psi}_t^j v_{x_t}^j, \frac{\dot{\psi}_t^j}{v_{x_t}^j}, \delta_t^j \Big], ~~\in \rr^{4} \\
\gamma_{3,t}^j &= \Big[\frac{\dot{\psi}_{t}^j}{v_{x_t}^j} , \frac{v_{y_t}^j}{v_{x_t}^j}, \delta_t^j \Big], ~~~~~~~~~~ \in \rr^{3}.
\end{align}
\end{subequations}
Note that the features are chosen based on the dynamic bicycle model, for further details we refer to \cite{SetMembership}.
\bibliographystyle{IEEEtran} 
\bibliography{IEEEabrv,mybibfile}

\end{document}